\documentclass{article}

\usepackage{arxiv}
\usepackage[utf8]{inputenc}
\usepackage[T1]{fontenc}
\usepackage[round]{natbib}
\renewcommand{\shorttitle}{MythoSemantic}
\makeatletter
\g@addto@macro\@floatboxreset{\def\@tempa{table}%
  \ifx\@captype\@tempa\columnwidth=0.6\textwidth\else\columnwidth=0.7\textwidth\fi}
\makeatother

\usepackage{microtype}
\usepackage{graphicx}
\usepackage{booktabs}
\usepackage{hyperref}
\usepackage{longtable}
\usepackage{array}

\usepackage{booktabs}
\usepackage{pdflscape}
\usepackage{adjustbox}
\usepackage{array}
\usepackage{xurl}
\usepackage{float}
\usepackage{xcolor}

\newcommand{\new}[1]{#1}
\newcommand{\pairsep}{\,\ensuremath{\leftrightarrow}\,}

\usepackage[english]{babel}

\title{MythoSemantic: An Interactive System for Discovering Semantic Parallels Across Mythological and Religious Traditions}

\author{
  \textbf{Nikita Ganziuk \textsuperscript{1}},
  \textbf{Innokentiy S. Humonen\textsuperscript{2, 3}},
  \textbf{Ilya Makarov\textsuperscript{2, 3, 4}},
    \textbf{Pavel Lebedev \textsuperscript{5}}  
\\[0.5em]
  \textsuperscript{1}HSE University,\quad
  \textsuperscript{2}AXXX,\quad
  \textsuperscript{3} Trusted AI Research Center, RAS \\ 
  \quad
  \textsuperscript{4} AI Talent Hub, ITMO University\quad
  \textsuperscript{5} Independent researcher\quad
\\[0.5em]
  \small{
    \textbf{Correspondence:} \href{mailto:haafus@gmail.com}{haafus@gmail.com}
  }
}

\begin{document}

\maketitle
\begin{abstract}
We present a tool for the comparative study of mythological and religious texts. Traditional manual analysis is highly labor-intensive and difficult to scale. To address this, we use neural language models to embed texts into a shared semantic space, \new{so that users can find semantic parallels across cultural traditions in an interactive interface.} \new{Embedded fragments form distinct semantic structures whose coarse geometry aligns with well-documented historical relations between traditions. We quantify how much of that geometry is already explained by lexical overlap and by proper names.}
\new{The current system is available at \url{https://mythoscope.io},} and the installable package is available at \url{https://github.com/haafus/MythoSemantic/blob/main/RUN.md}.
\end{abstract}

\section{Introduction}
\begin{figure*}[ht]
\centering
\includegraphics[width=0.94\textwidth]{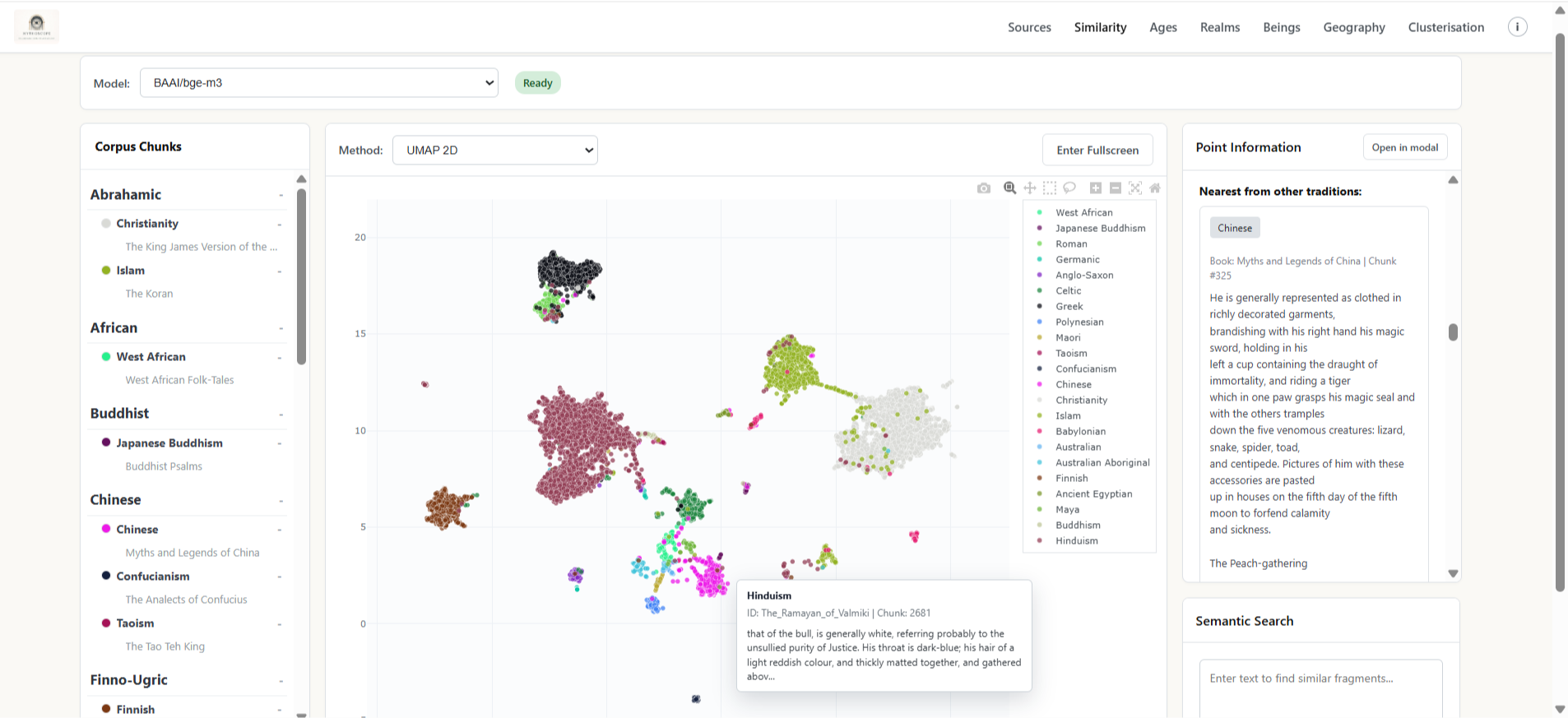}
\caption{The \textit{Similarity} view of the MythoSemantic interface. The left panel provides a tradition-grouped browser of the corpus; the central panel shows a 2D UMAP projection of BAAI/bge-m3 embeddings, with a tooltip surfacing the tradition, source text, and passage for the fragment under the cursor; the right panel displays the nearest cross-tradition neighbor of the currently selected fragment alongside a free-text semantic search box.}
\label{fig:interface-similarity}
\end{figure*}

\begin{figure*}[h!]
\centering
\includegraphics[width=0.88\textwidth]{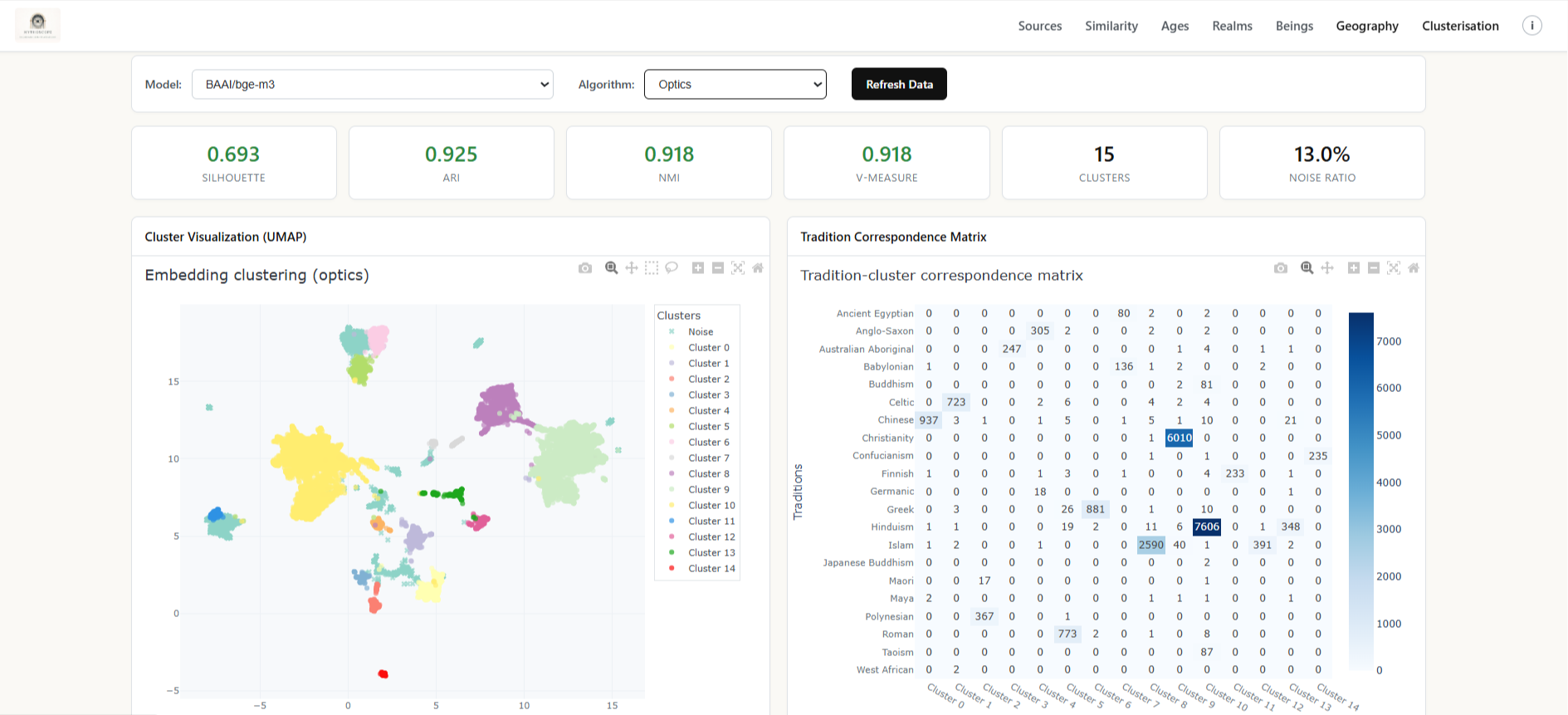}
\caption{The \textit{Clusterisation} view: quality metrics (silhouette, ARI, NMI, V-measure, cluster count, noise ratio), a UMAP cluster visualization, and the tradition-cluster correspondence matrix, shown here for BAAI/bge-m3 with OPTICS.}
\label{fig:interface-clusterisation}
\end{figure*}

The comparative analysis of mythological narratives and motifs remains a predominantly manual process. \new{It needs specialists, who are few, takes a great deal of labor, and does not scale to large corpora.} \new{Neural language models now make it possible to search such texts for semantic parallels and similar motifs across traditions.} \new{An automated tool that shows semantic relationships among such texts is therefore useful for comparative mythology and digital humanities research.} \new{Such a system lets one scholar scan a corpus that would take a career to read by hand, and lets the same query be repeated whenever the corpus grows or the embedding model is replaced.} While the term ``motif'' is vague, the \new{system} (\autoref{fig:interface-similarity}, \autoref{fig:interface-clusterisation}) \new{is} a guiding map for deeper research by anthropologists and ethnologists.

\section{Related Work}
Historically, the comparative analysis of mythological and religious narratives has relied on the manual extraction and cross-referencing of motifs across distinct cultural traditions \citep{thompson1955motif, uther2004types}. Structuralist approaches, such as those by \citet{propp1968morphology}, demonstrated the possibility of identifying universal structural elements in diverse narratives. \new{These qualitative methods are labor-intensive, depend on subjective interpretation, and do not scale to fragmented global corpora.}

The development of distributional semantics \citep{mikolov2013distributed}
initiated the transition from traditional lexical matching to word
embeddings. This trajectory continued with the introduction
of the attention mechanism and the Transformer architecture
\citep{vaswani2017attention}, \new{which made representations depend on context and led to large
language models} and their corresponding text-embedding counterparts, such as the
Qwen3 Embedding series \citep{zhang2025qwen3embedding}. \new{Such models turn whole passages into embeddings that capture much of
their meaning.}

Consequently, passages with similar meanings are mapped to nearby regions of the embedding space, even when they differ lexically. The synthesis of computational methods and humanities research led to the concept of ``distant reading'' \citep{moretti2013distant}, shifting the focus from the close reading of individual texts to the macro-analysis of large corpora. Within the framework of ``computational folkloristics'' \citep{abello2012computational}, \new{researchers have tried to find connections between texts algorithmically.}
Several recent NLP systems address closely related sub-problems, targeting
motif and narrative similarity within modern texts, single tale types, or
individual traditions \citep{yarlott2024golem, hatzel2024storyembeddings,
hatzel2024tellmeagain, arcon2025cinderella, alyami2026arabiannights,
hatzel2026semeval}. None of these systems, however, is interactive or
supports open-ended, cross-tradition exploration by non-programmer users\new{;
MythoSemantic fills this gap.}
\new{Multilingual embeddings have rarely been used to map belief systems across the world.} \new{Such a study needs a browsable map: the whole corpus laid out in one space, every fragment clickable, and every retrieved neighbor traceable back to its source text and tradition.}

\section{Data}
\label{sec:data}
The dataset comprises 28 mythological and religious texts, representing 22 traditions across 12 major cultural areas, all provided in English translations and derived from open sources, mainly \textit{Project Gutenberg}, with source-specific boilerplate removed by a dedicated cleaning script. \autoref{tab:corpus-traditions} \new{in the appendix} lists the traditions and their corresponding source text(s); several traditions (e.g.\ Hinduism, Babylonian, Greek) are represented by multiple texts, while others are represented by a single one, so the corpus is imbalanced across traditions\new{, and larger traditions form correspondingly larger regions in the learned semantic space. A single language of translation makes the traditions directly comparable; \S\ref{sec:controls} measures how much of the resulting structure comes from shared vocabulary.} \new{We compare the semantic structure the system finds with the tradition label of each text} (\autoref{sec:evaluation}).

Instead of embedding complete documents, the corpus is first divided into overlapping fragments. Although such fragments should not be equated with motifs, they provide a practical approximation of this abstract concept by representing relatively self-contained semantic units that can be compared computationally across texts. This approach also satisfies the input length constraints of embedding models while reducing the risk that multiple independent storylines within a single source are merged into a single representation. \new{The chunking procedure is described in \autoref{sec:chunking}.} Each resulting chunk is stored with a sequential index and its source metadata, yielding \new{25{,}671} overlapping fragments across the corpus.

\section{System Overview}
\new{MythoSemantic is built for digital humanities researchers, comparative-religion and folklore scholars, and students who explore semantic relationships across mythological corpora without programming.} \new{We turn the texts into vectors and hand the resulting structure back to a human reader to interpret.} 

\paragraph{Minimum system requirements.} The server requires Python~3.9 or later and runs on Linux, macOS, and Windows; the interface is accessed through any modern web browser. No GPU is required: precomputed vector databases for all five embedding models are downloaded automatically on first launch, so a standard laptop with a multi-core CPU, 8~GB of RAM and approximately 2~GB of free disk space is sufficient for the default corpus. An internet connection is needed only for installation and the initial download; all subsequent analysis runs fully locally.

\subsection{Pipeline}
Source texts are ingested, converted to a unified format, and organized by cultural-historical tradition (\textbf{Loader}). Documents are then segmented into overlapping fragments, as individual sources typically interleave multiple distinct storylines, and each fragment is embedded using one of the five embedding models listed in \autoref{tab:embedding_grid}; the resulting vectors are stored for fast nearest-neighbor retrieval, enabling both point-based and free-text similarity queries (\textbf{Embeddings}). The precise definition of a ``motif'' remains contested in the folkloristics literature \citep{thompson1955motif}, and recent benchmarks show that embedding size alone is a poor predictor of narrative-similarity quality \citep{hatzel2026semeval}; we therefore evaluate five embedding models spanning different architectures, training eras, and parameter scales, rather than assuming a single optimal embedding model. For exploratory analysis, embeddings are projected to two dimensions via PCA, t-SNE, and UMAP, and an inter-tradition distance heatmap is computed from averaged per-tradition vectors (\textbf{Visualization}). Finally, seven clustering algorithms, namely KMeans \citep{macqueen1967kmeans}, HDBSCAN \citep{campello2013hdbscan}, OPTICS \citep{ankerst1999optics}, Spectral Clustering \citep{ng2002spectral}, Gaussian Mixture Models \citep{dempster1977em}, BIRCH \citep{zhang1996birch}, and MeanShift \citep{comaniciu2002meanshift} group fragments in the semantic space independently of any predefined schema, and the resulting clusters are compared against tradition labels to reveal both expected and unforeseen cross-cultural connections (\textbf{Clustering}).

\begin{table}[htbp]
\centering
\resizebox{\columnwidth}{!}{%
\begin{tabular}{lc}
\toprule
\textbf{Model} & \textbf{Vector Dimension} \\
\midrule
\begin{tabular}[c]{@{}l@{}}sentence-transformers/\\ paraphrase-multilingual-MiniLM-\\ L12-v2\end{tabular} & 384 \\
\begin{tabular}[c]{@{}l@{}}sentence-transformers/\\ LaBSE\end{tabular} & 768 \\
intfloat/e5-large-v2 & 1024 \\
BAAI/bge-m3 & 1024 \\
Qwen/Qwen3-Embedding-4B & 2560 \\
\bottomrule
\end{tabular}%
}
\caption{Embedding models and their respective vector dimensions.}
\label{tab:embedding_grid}
\end{table}

\subsection{Interface}
The web interface \new{brings the pipeline's outputs together in four linked views, so scholars can interpret them without writing code.} The \textit{Sources} view provides a tradition-grouped browser of the corpus, a reading pane for individual texts, and per-document metadata with export options. The \textit{Similarity} view is the primary analytical surface: it displays the two-dimensional projection for a chosen embedding model, where clicking any point surfaces its nearest neighbors across the corpus, and offers a free-text search box that embeds an arbitrary user query and retrieves the closest matching fragments regardless of tradition (\autoref{fig:interface-similarity}). The \textit{Geography} view situates traditions on an interactive map, linking each region to its associated source texts. The \textit{Clusterisation} view exposes the quantitative results of \autoref{sec:evaluation} interactively: selecting an embedding model and clustering algorithm returns the corresponding quality metrics alongside a colored projection and a cluster-to-tradition confusion matrix (\autoref{fig:interface-clusterisation}).

For example, a researcher browsing the \textit{Ramayana of Valmiki} (Hinduism) in \textit{Sources} can switch to \textit{Similarity} and click on a passage describing a deity's bull, throat dark-blue and hair reddish; the nearest cross-tradition neighbor surfaced in the right-hand panel is a passage from \textit{Myths and Legends of China} (Chinese tradition) describing a deity clothed in richly decorated garments who holds a cup containing the elixir of immortality and rides a tiger: a parallel the researcher can inspect, compare, and follow up on without ever leaving the browser or engaging with the underlying embedding models.

\new{Each answer the system gives is a passage from the corpus. A retrieved parallel arrives with its tradition, its source text and its position in that text, and the free-text box lets a scholar ask for one in ordinary language instead of in code. We keep generation out of this path deliberately, because a fluent gloss of why two passages match is easy to believe and hard to check, and in comparative mythology such a claim is expensive to retract once it has been repeated. The system reports what a scholar can verify against the text itself, namely the provenance of every fragment, the cluster-to-tradition matrix, and the fragments that sit outside their own tradition's cluster.}

\section{Evaluation}
\label{sec:evaluation}

We manually reviewed the two-dimensional projections and distance heatmaps for all five embedding models (\autoref{fig:all_models} \new{in the appendix}) and selected BGE-M3 as a representative example for the discussion below, \new{because it separates traditions clearly while keeping closely related groups of texts near each other.}

\subsection{Qualitative Validation}
\label{sec:qualitative-validation}

\begin{figure}[t]
\centering
\includegraphics[width=\columnwidth]{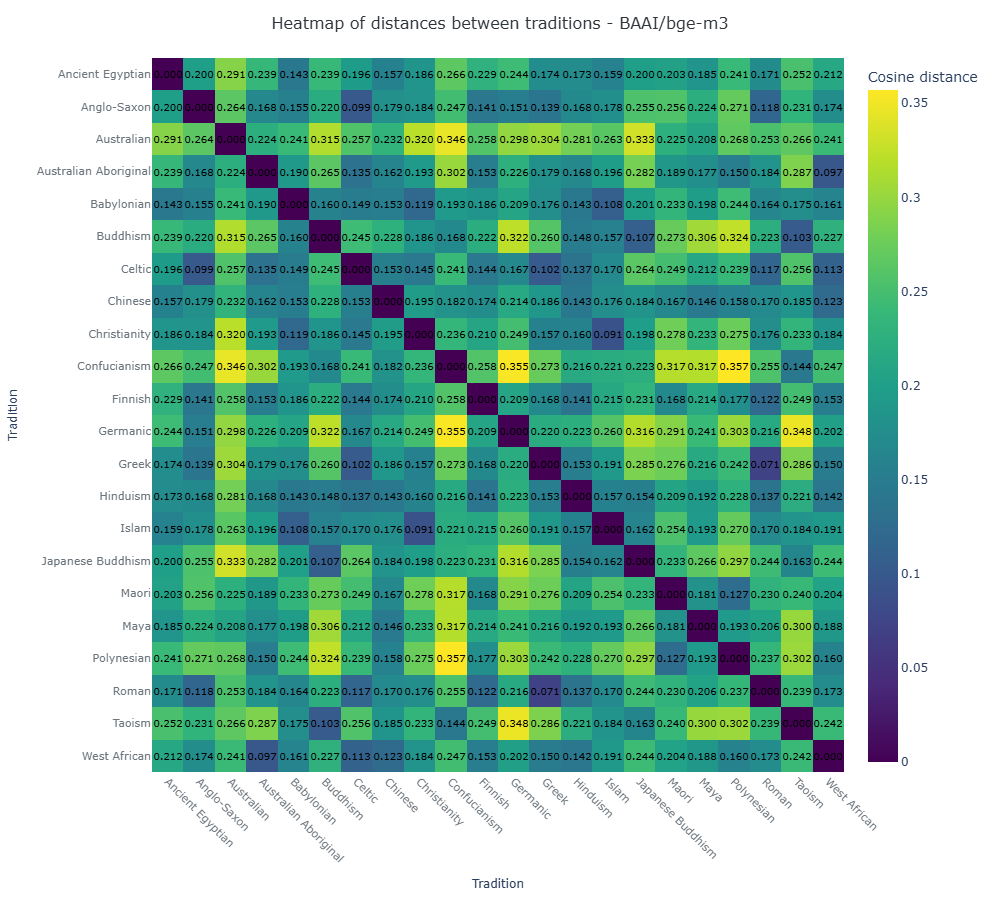}
\caption{Distance heatmap between traditions for BAAI/bge-m3, as rendered in the \textit{Similarity} view (Method: Distance Heatmap).}
\label{fig:bge_heatmap}
\end{figure}

An analysis of the distance heatmap (\autoref{fig:bge_heatmap}) reveals several \new{clear} groups of closely related traditions. The closest pair in the corpus is Greek and Roman, with a distance of 0.071, as expected given the deep historical relationship between the two traditions. \new{\S\ref{sec:controls} measures how much of this proximity comes from the vocabulary of the English translations.} The second distinct group consists of Abrahamic and Middle Eastern texts. The distance between Christianity and Islam is 0.091, between Babylonian and Islam is 0.108, and between Babylonian and Christianity is 0.119. Such proximity may reflect the shared religious and cultural environment of the Middle East. The third group is formed by East Asian and Buddhist traditions. Buddhism and Taoism appear closely related with a distance of 0.103, Buddhism and Japanese Buddhism with a distance of 0.107, as well as Confucianism and Taoism with 0.144. These traditions are united by the philosophical and religious nature of their texts\new{ and by a shared aphoristic, didactic register}.

A separate European block is formed by Anglo-Saxon, Celtic, and Roman. The distance between Anglo-Saxon and Celtic is 0.099, between Celtic and Roman is 0.117, and between Anglo-Saxon and Roman is 0.118. An unexpected result is the proximity of Australian Aboriginal and West African, with a distance of 0.097. The proximity of Polynesian and Maori is also evident, with a distance of 0.127, as is that of Polynesian and West African, at 0.160. In the case of Maori and Polynesian, the result is expected, as these traditions are culturally connected. \new{The Australian Aboriginal\pairsep{}West African pair, by contrast, corresponds to no documented contact; the system surfaces such pairs as leads for a philologist to follow.}

\subsection{\new{What the Geometry Is Made Of}}
\label{sec:controls}
\new{A proximity between two traditions in \autoref{fig:bge_heatmap} may reflect shared
narrative content, or only the vocabulary and onomastics of their English translations.
A lexical control separates the two. Over the same 25{,}671 fragments we fit a TF-IDF
model and derive a tradition-by-tradition distance matrix by the same construction used
for the embedding heatmap. We then repeat it twice, once with proper names removed and
once with proper names alone, counting as a proper name any token capitalized in at least
85\% of its non-sentence-initial occurrences (3{,}814 types).}

\new{Across the 231 tradition pairs the lexical matrix correlates with the BGE-M3 matrix at
Spearman $\rho = 0.76$ (Mantel permutation test, $p < 10^{-3}$), leaving 42\% of the rank
variance unexplained by word overlap. The structure does not come from proper names: the
names-only matrix reaches $\rho = 0.40$, clustering fragments by names alone recovers
traditions at ARI $0.09$, and deleting every name from the corpus moves the lexical matrix
by $\Delta\rho = 0.011$.}

\begin{table}[t]
\centering
\small
\resizebox{\columnwidth}{!}{%
\begin{tabular}{lrrrr}
\toprule
& \multicolumn{4}{c}{\new{Rank among 231 pairs}} \\
\cmidrule(l){2-5}
\new{Tradition pair} & \new{Emb.} & \new{Lex.} & \new{No names} & \new{Names} \\
\midrule
\new{Greek\pairsep{}Roman}            & \new{1}  & \new{9}  & \new{17} & \new{5} \\
\new{Christianity\pairsep{}Islam}     & \new{2}  & \new{1}  & \new{5}  & \new{1} \\
\new{Austr.\ Abor.\pairsep{}W.\ African} & \new{3} & \new{8} & \new{10} & \new{131} \\
\new{Anglo-Saxon\pairsep{}Celtic}     & \new{4}  & \new{73} & \new{53} & \new{90} \\
\new{Buddhism\pairsep{}Taoism}        & \new{6}  & \new{27} & \new{31} & \new{223} \\
\new{Babylonian\pairsep{}Islam}       & \new{8}  & \new{43} & \new{76} & \new{3} \\
\new{Polynesian\pairsep{}Maori}       & \new{15} & \new{56} & \new{63} & \new{10} \\
\bottomrule
\end{tabular}%
}
\caption{\new{The tradition pairs discussed in \S\ref{sec:qualitative-validation}, ranked by
closeness under the BGE-M3 embedding and under three TF-IDF views of the same fragments:
full text, text with proper names removed, and proper names only. A low embedding rank next
to a high lexical rank marks a proximity that word overlap does not explain.}}
\label{tab:controls}
\end{table}

\new{The control also works lead by lead (\autoref{tab:controls}). Greek\pairsep{}Roman,
Anglo-Saxon\pairsep{}Celtic and Buddhism\pairsep{}Taoism rank higher under the embedding than
under any lexical view, the last two by a wide margin, so shared words do not explain them.
Christianity\pairsep{}Islam is the closest pair both lexically and by proper names alone, and
Babylonian\pairsep{}Islam ranks third by names alone; a scholar should read both as partly
onomastic. The Australian Aboriginal\pairsep{}West African proximity survives the removal of
names but is visible to the lexical model as well, which points to a shared folk-tale register
rather than to contact.}

\subsection{Quantitative Clustering Results}

Across different embedding models and clustering algorithms, unsupervised clusters recover the original tradition labels with high fidelity: density-based methods reach ARI values of 0.90--0.98 (HDBSCAN/OPTICS on BGE-M3 and E5-large-v2), and even the weaker fixed-$k$ algorithms (KMeans, GMM, BIRCH) still recover a coherent, non-random partial structure (ARI $\approx$ 0.3--0.5) despite being forced into a suboptimal number of clusters. This consistency across five architecturally distinct embedding models and seven clustering algorithms indicates that the semantic space itself is well-formed, rather than the result of a single fortunate model--algorithm combination.

\begin{table}[t]
\centering
\small
\resizebox{\columnwidth}{!}{%
\begin{tabular}{lrrrrrr}
\toprule
Algorithm & Sil. & ARI & NMI & V & Clusters & Noise \\
\midrule
BIRCH             & 0.019 & 0.410 & 0.690 & 0.690 & 22 & 0.0\% \\
Gaussian Mixture  & 0.472 & 0.463 & 0.770 & 0.770 & 22 & 0.0\% \\
HDBSCAN           & 0.688 & 0.904 & 0.908 & 0.908 & 15 & 3.1\% \\
K-means           & 0.033 & 0.320 & 0.651 & 0.651 & 22 & 0.0\% \\
MeanShift         & 0.573 & 0.823 & 0.783 & 0.783 &  7 & 0.0\% \\
OPTICS            & 0.693 & 0.925 & 0.918 & 0.918 & 15 & 13.0\% \\
Spectral          & 0.032 & 0.841 & 0.849 & 0.849 & 22 & 0.0\% \\
\bottomrule
\end{tabular}%
}
\caption{Clustering metrics for \texttt{BAAI/bge-m3}. Sil. = silhouette coefficient; ARI = Adjusted Rand Index against tradition labels; NMI = Normalized Mutual Information; V = V-measure; Clusters and Noise = number of clusters discovered and the percentage of fragments left unassigned as noise (applicable to density-based algorithms only).}
\label{tab:clustering-bge-m3}
\end{table}

Two aspects of this structure are, however, more informative than the raw scores. First, the number of clusters (see \autoref{tab:clustering-bge-m3}) that density-based algorithms discover is consistently lower than the nominal number of traditions, 15 for HDBSCAN and OPTICS, and as few as 7 under MeanShift, indicating that several traditions are not semantically distinct at this level of granularity and instead merge into shared superclusters (cf.\ the Greco-Roman and Abrahamic groupings in \S\ref{sec:qualitative-validation}). Second, even the best-performing configurations leave a non-trivial fraction of fragments unassigned relative to their nominal tradition (3.1\% noise for HDBSCAN and 13.0\% for OPTICS on BGE-M3, \autoref{tab:clustering-bge-m3}). \new{Each such outlier is a fragment whose content diverges from its own tradition's cluster, which makes it a natural candidate for a cross-tradition parallel and, in our view, the most useful output of the pipeline for domain experts.} The full per-model, per-algorithm breakdown for the remaining four embedding models is provided in the Appendix (Tables~\ref{tab:clustering-qwen3}--\ref{tab:clustering-minilm}).


MiniLM-L12-v2 is a clear exception to this pattern\new{: at 384 dimensions the smallest backbone evaluated, it} fails to form usable structure altogether (HDBSCAN: 3 clusters, ARI 0.019, see \autoref{tab:clustering-minilm}), \new{which suggests} that a minimum representational capacity is necessary.

\section{Availability and Licensing}
MythoSemantic is publicly available at \url{https://github.com/haafus/MythoSemantic} under the MIT license, covering the corpus-collection, embedding, and clustering pipeline as well as the web interface described above; setup and running instructions are provided in the repository's \href{https://github.com/haafus/MythoSemantic/blob/main/RUN.md}{\texttt{RUN.md}}.

\section{Conclusion}
We have presented MythoSemantic, an interactive platform for the computational comparative analysis of mythological texts, built on the premise that modern natural language processing methods can complement the classical objectives of comparative mythology. \new{Our analysis shows that mythological and religious texts form distinct, semantically coherent regions within the embedding space, and that the coarse geometry of those regions is consistent with well-documented cross-tradition relations while remaining only partly reducible to lexical overlap. The system therefore hands the scholar a ranked set of candidates: neighboring fragments, adjacent traditions, and fragments that sit outside their own tradition's cluster. Philological work decides whether any of them is a genuine parallel.}

\section{Future Work}
\new{The controls of \S\ref{sec:controls} measure how much of the embedding geometry comes from vocabulary. Future work will separate narrative structure from style further} through explicit character- and relationship-graph representations, continual adoption of newer embedding backbones as they emerge, corpus expansion, and linking discovered clusters to established motif indices such as Thompson's Motif-Index and the Aarne--Thompson--Uther catalog \citep{thompson1955motif, uther2004types}. The interface additionally includes experimental graph-based \textit{Ages}, \textit{Realms}, and \textit{Beings} views for browsing the corpus by temporal, spatial, and entity-based facets respectively.
\new{The latest version of MythoSemantic is available at \url{https://mythoscope.io}.}

\bibliographystyle{acl_natbib}
\bibliography{custom}

\appendix
\label{sec:appendix}
\renewcommand{\thefigure}{\thesection-\arabic{figure}}
\counterwithin{figure}{section} 
\renewcommand{\thetable}{\thesection-\arabic{table}}
\counterwithin{table}{section} 

\newpage


\section{Corpus}

\begin{table}[h]
\centering
\resizebox{\columnwidth}{!}{%
\begin{tabular}{ll}
\toprule
Tradition & Source text(s) \\
\midrule
Christianity & King James Bible \\
Islam & Quran \\
Confucianism & Analects of Confucius \\
Hinduism & Upanishads; Ramayana of Valmiki; \\
 & Bhagavad Gita; Mahabharata \\
Buddhism & Dhammapada \\
Taoism & Tao Te Ching \\
Chinese & Myths and Legends of China \\
Japanese Buddhism & Buddhist Psalms \\
Ancient Egyptian & Book of the Dead \\
Babylonian & Babylonian Legends of Creation; \\
 & Code of Hammurabi \\
Greek & Iliad; Odyssey \\
Roman & Aeneid \\
Germanic & Nibelungenlied; Poetic (Elder) Edda \\
Anglo-Saxon & Beowulf \\
Celtic & Mabinogion \\
Finnish & Kalevala \\
West African & West African Folk-Tales \\
Australian & Australian Legendary Tales \\
Australian Aboriginal & Myths of the Australian Aborigines \\
Maori & Te Tohunga \\
Polynesian & Legends of Maui \\
Maya & Popol Vuh \\
\bottomrule
\end{tabular}%
}
\caption{The 22 traditions in the corpus and their source text(s).}
\label{tab:corpus-traditions}
\end{table}

\section{Chunking}
\label{sec:chunking}

\new{We chunk documents recursively to keep each fragment semantically coherent. Documents are first split at paragraph boundaries, with adjacent paragraphs merged until a configurable length limit is reached. If a paragraph exceeds this limit, it is recursively divided into sentences; oversized sentences are then split at punctuation boundaries, and only as a last resort at the word or character level. This order of separators keeps paragraphs and sentences intact whenever possible. To maintain continuity across chunk boundaries, a configurable overlap is introduced between adjacent fragments, with both sentence-based and character-based overlap strategies implemented.}

\newpage
\section{Clustering tables}

\begin{table}[h]
\centering
\resizebox{\columnwidth}{!}{%
\begin{tabular}{lrrrrr}
\toprule
Algorithm & Sil. & ARI & NMI & V & Clusters/Noise \\
\midrule
BIRCH             & 0.033 & 0.332 & 0.662 & 0.662 & 22 / 0.0\% \\
Gaussian Mixture  & 0.525 & 0.497 & 0.748 & 0.748 & 22 / 0.0\% \\
HDBSCAN           & 0.633 & 0.764 & 0.850 & 0.850 & 14 / 2.4\% \\
K-means           & 0.042 & 0.306 & 0.648 & 0.648 & 22 / 0.0\% \\
MeanShift         & 0.566 & 0.797 & 0.782 & 0.782 &  7 / 0.0\% \\
OPTICS            & 0.731 & 0.878 & 0.888 & 0.888 & 14 / 33.6\% \\
Spectral          & 0.041 & 0.791 & 0.830 & 0.830 & 22 / 0.0\% \\
\bottomrule
\end{tabular}%
}
\caption{Clustering metrics for \texttt{Qwen/Qwen3-Embedding-4B}. Sil. = silhouette coefficient; ARI = Adjusted Rand Index
against tradition labels; NMI = Normalized Mutual Information; V = V-measure; Clusters/Noise =
number of clusters discovered and the percentage of fragments left unassigned as noise (applicable to
density-based algorithms only).}
\label{tab:clustering-qwen3}
\end{table}

\begin{table}[h]
\centering
\resizebox{\columnwidth}{!}{%
\begin{tabular}{lrrrrr}
\toprule
Algorithm & Sil. & ARI & NMI & V & Clusters/Noise \\
\midrule
BIRCH             & --- & --- & --- & --- &  1 / 0.0\% \\
Gaussian Mixture  & 0.507 & 0.498 & 0.807 & 0.807 & 22 / 0.0\% \\
HDBSCAN           & 0.699 & 0.977 & 0.960 & 0.960 & 16 / 3.4\% \\
K-means           & 0.031 & 0.368 & 0.701 & 0.701 & 22 / 0.0\% \\
MeanShift         & 0.394 & 0.564 & 0.666 & 0.666 &  6 / 0.0\% \\
OPTICS            & 0.687 & 0.973 & 0.956 & 0.956 & 16 / 1.8\% \\
Spectral          & 0.037 & 0.887 & 0.889 & 0.889 & 22 / 0.0\% \\
\bottomrule
\end{tabular}%
}
\caption{Clustering metrics for \texttt{intfloat/e5-large-v2}. Sil. = silhouette coefficient; ARI = Adjusted Rand Index
against tradition labels; NMI = Normalized Mutual Information; V = V-measure; Clusters/Noise =
number of clusters discovered and the percentage of fragments left unassigned as noise (applicable to
density-based algorithms only).}
\label{tab:clustering-e5-large}
\end{table}

\vfill\eject

\begin{table}[h]
\centering
\resizebox{\columnwidth}{!}{%
\begin{tabular}{lrrrrr}
\toprule
Algorithm & Sil. & ARI & NMI & V & Clusters/Noise \\
\midrule
BIRCH             & 0.024 & 0.402 & 0.647 & 0.647 & 22 / 0.0\% \\
Gaussian Mixture  & 0.436 & 0.471 & 0.737 & 0.737 & 22 / 0.0\% \\
HDBSCAN           & 0.616 & 0.716 & 0.793 & 0.793 & 15 / 4.8\% \\
K-means           & 0.029 & 0.287 & 0.592 & 0.592 & 22 / 0.0\% \\
MeanShift         & 0.615 & 0.688 & 0.755 & 0.755 & 11 / 0.0\% \\
OPTICS            & 0.801 & 0.906 & 0.906 & 0.906 & 13 / 40.0\% \\
Spectral          & 0.020 & 0.731 & 0.734 & 0.734 & 22 / 0.0\% \\
\bottomrule
\end{tabular}%
}
\caption{Clustering metrics for \texttt{sentence-transformers/LaBSE}. Sil. = silhouette coefficient; ARI = Adjusted Rand Index
against tradition labels; NMI = Normalized Mutual Information; V = V-measure; Clusters/Noise =
number of clusters discovered and the percentage of fragments left unassigned as noise (applicable to
density-based algorithms only).}
\label{tab:clustering-labse}
\end{table}

\begin{table}[h]
\centering
\resizebox{\columnwidth}{!}{%
\begin{tabular}{lrrrrr}
\toprule
Algorithm & Sil. & ARI & NMI & V & Clusters/Noise \\
\midrule
BIRCH             &  0.006 & 0.247 & 0.455 & 0.455 & 22 / 0.0\% \\
Gaussian Mixture  &  0.312 & 0.334 & 0.569 & 0.569 & 22 / 0.0\% \\
HDBSCAN           &  0.102 & 0.019 & 0.088 & 0.088 &  3 / 2.4\% \\
K-means           &  0.026 & 0.221 & 0.451 & 0.451 & 22 / 0.0\% \\
MeanShift         &  0.429 & 0.533 & 0.540 & 0.540 &  4 / 0.0\% \\
OPTICS            &  0.099 & 0.018 & 0.086 & 0.086 &  3 / 1.1\% \\
Spectral          & -0.003 & 0.450 & 0.540 & 0.540 & 22 / 0.0\% \\
\bottomrule
\end{tabular}%
}
\caption{Clustering metrics for \texttt{MiniLM-L12-v2}. Sil. = silhouette coefficient; ARI = Adjusted Rand Index
against tradition labels; NMI = Normalized Mutual Information; V = V-measure; Clusters/Noise =
number of clusters discovered and the percentage of fragments left unassigned as noise (applicable to
density-based algorithms only).}
\label{tab:clustering-minilm}
\end{table}

\newpage
\section{Projections for all models}

\begin{figure}[h]
\centering
\includegraphics[width=\columnwidth]{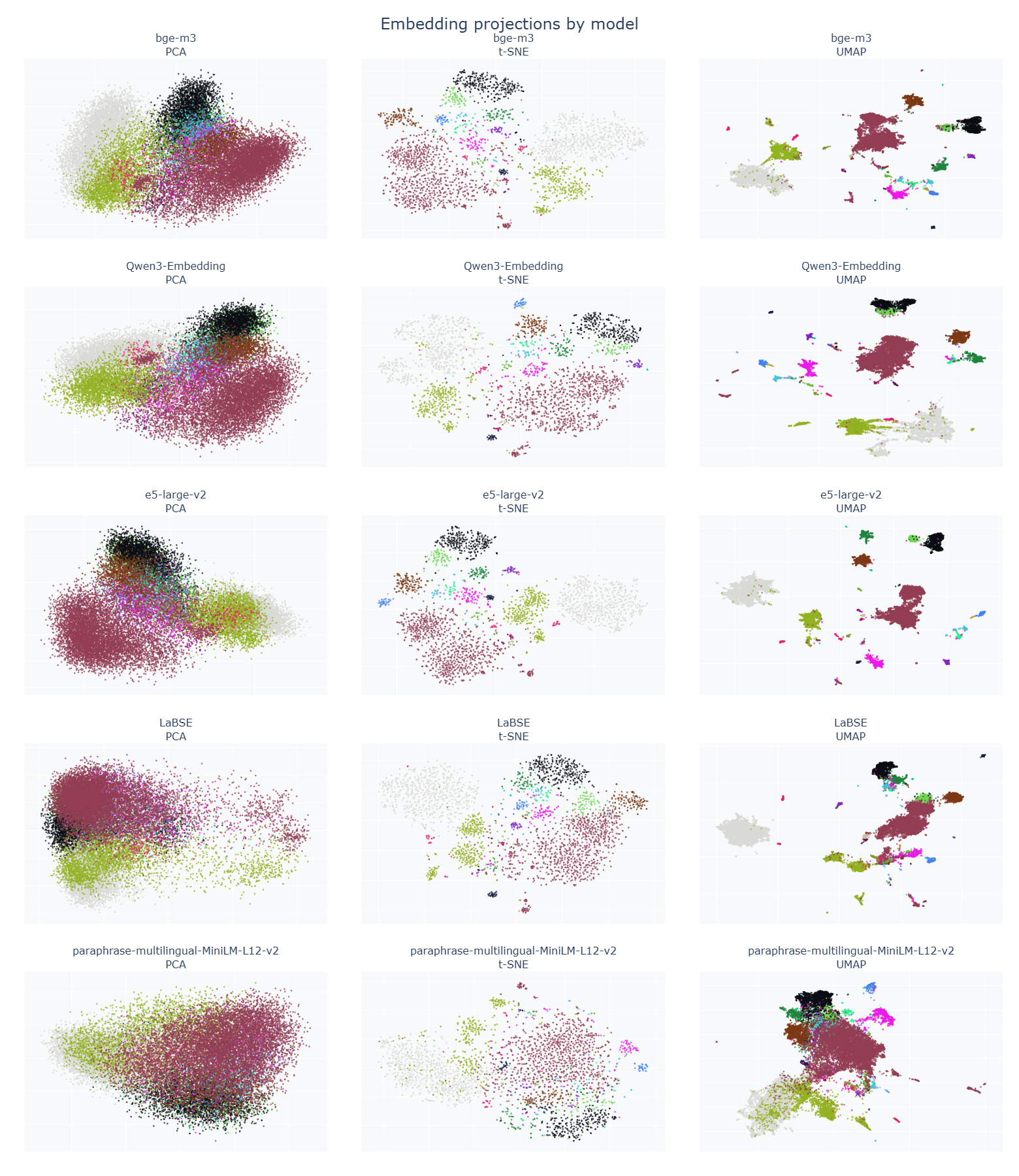}
\caption{Visualizations for all used models and dimension reduction algorithms.}
\label{fig:all_models}
\end{figure}

\end{document}